# Predicting opponent team activity in a RoboCup environment

**By Selene Báez Santamaría**
**Supervised by Jim Little**

University of British Columbia

**Abstract**

The goal of this project is to predict the opponent's configuration in a RoboCup SSL environment. For simplicity, a Markov model assumption is made such that the predicted formation of the opponent team only depends on its current formation. The field is divided into a grid and a robot state per player is created with information about its position and its velocity. To gather a more general sense of what the opposing team is doing, the state also incorporates the team's average position (centroid). All possible state transitions are stored in a hash table that requires minimum storage space. The table is populated with transition probabilities that are learned by reading vision packages and counting the state transitions regardless of the specific robot player. Therefore, the computation during the game is reduced to interpreting a given vision package to assign each player to a state, and looking for the most likely state it will transition to. The confidence of the predicted team's formation is the product of each individual player's probability. The project is noteworthy in that it minimizes the time and space complexity requirements for opponent's moves prediction.

## 1. Introduction

RoboCup is an international robotics competition primarily focused on soccer tournaments. Since its beginnings in 1997, it has been a motivation for Artificial Intelligence (AI) and robotics research in various universities around the world because of the challenges it raises[1]. The tournament has several categories, one of which is the Small Size League (SSL). In this category each team consists of six cylinder-shaped robots and a golf ball that are tracked by a vision system. The vision system consists of two overhead cameras that distribute the information through vision packages to both teams for coaching and tactic planning purposes

### 1.1. Motivation

As mentioned, soccer raises several challenges from an AI perspective. It is played on a highly dynamic environment with several players and one ball moving simultaneously. Each robot must have a cooperative element to work with its teammates, while simultaneously having a competitive element to beat the opponent. Moreover, their environment is not fully observable since the intentions and tactics of the opponent team are not visible to a team.

These challenges must be overcome to achieve one of the most desired advantages in soccer: responsiveness. A responsive team is able to adapt its decisions (e.g. its defence decisions) to the changes in the environment (e.g. type of offence the opponent brings to the game). If it is able to predict the play, or sequence of movements the players from the opponent team perform over a period of time, it can block and/or intercept passes, or prevent players from moving to potentially dangerous areas and thus interrupt the offence play.

However, predicting plays is a very complex decision-making process to model because they are dependent on many factors. In an SSL environment, the opponent team will most likely have a repertory of predesigned plays and a general strategy it tries to follow. Yet, while both the repertory and the general strategy are necessary for the opponent to choose and perform an offence play, in a real-life situation these are not sufficient. Some plays, even when carefully planned, cannot be correctly performed because of other factors that change or end the predesigned play. Hence, just like the best defence, the best offence will adapt to the way the match develops.

It follows that, while predicting a predesigned play only requires learning the opponent's repertory and finding out their strategy, if a team wants to be prepared for a more realistic offence, reading the match constantly is more valuable.

This work proposes that the flow of offence adaptations can be simplified from predicting predesigned plays to predicting a team's formation which only captures the configuration of the players at a given time. This way, the team would still be able to intercept and block the opponent because it will know the most likely formation to come. However, the main difference is that predicting a play focuses on a longer period of time and leaves out possible changes or adaptations throughout the match or the play itself, while predicting formations allows for a more responsive defence that only foresees a short period of time.

## 2. Related work

### 2.1. Modeling the opponent

Considered as one of the best teams in the SSL, the CMDragons from Carnegie Mellon University were ranked second best in the 2014 RoboCup in Brazil. Some of their best work has focused on modeling the opponent[2,3] using probabilistic approaches[4] and aiming to predict plays[5]. In their last successful attempt, they used their characteristic method for dividing the match into *episodes*, and ran a clustering algorithm that recognizes the opponent's predesigned plays. This system gave them a great advantage in the 2011 tournament which reflected a stronger defence.

Even so, it must be said that this approach is outdated since it was implemented and tested three years ago. At that point, most of the participant teams showed pre-designed tactics with little to no adaptive plays. The CMDragons trained their system with games from previous tournaments, and their opponents' plays did not change significantly before the next tournament[6]. For this reason they were able to create a precise model and use a play-based approach. Yet, as teams become more aware of the need of responsiveness, a play approach is no longer the best option.

A different approach was taken by Peter Carr and his team[7]. They developed the idea of modeling the opponent team through occupancy maps, which is particularly useful for noisy data like the RoboCup vision system. The approach consists on dividing the field into a grid, and count the number of players per team, per bin. The benefit of using a grid is that if the readings from the vision system are too noisy, it would not affect an occupancy map as much.

Secondly, Carr explores the idea of team centroids. The team centroids incorporate information about the team as a whole by representing the average position with an "X" and the standard deviation around the position with a shaded area. This approach allows to get information about the team configuration by looking at central tendency measures.

### 2.2. Memory and running time bounds

Considerations about memory-bounded agents are important as Peter Stone exposed during his workshop in 2013[8]. An acceptable approach must have memory requirements that are low enough for a SSL RoboCup robot to store, and be efficient about the learned information. Moreover, responsiveness requires readings and interpretations to be done quickly (i.e. in a fraction of a second) to allow for the proper reactions to take place while they are still useful. Thus, running time is also an important concern[9] because the predictions the system aims to generate are in the near future and depend on constantly updated vision packages.

### 3. Markov model

This project models the opponent behaviour by making a Markov assumption. This way, the opponent's formation on the closest time step only depends on the formation it currently has. The simplification brings as consequence a more responsive defence that only worries about the closest point in time it can interrupt the opponent's offence.

The current and predicted formations relate by breaking the formation into individual moves per player. This means that the team configuration is interpreted by the vision system as different players being in and moving to different "states". As such, each robot is considered as an individual entity that will most likely move to a different state. The transitions are not dependent on the particular player, but on the state, where a state refers to cohesive information about the position and velocity of the individual robot, and the team tendency as a whole. This is more representative of how soccer is played, because coaches can replace players or change their "roles" in the game, however, no matter who is playing, the player follows a behaviour specified by how the team is playing and his function as part of the team.

### 3.1. States

Building on the ideas by Carr, a comprehensive state incorporates information about the team as an entity as well as the robot's individual information. Every time a vision package arrives, the team centroid is calculated and set for all the robots in the formation. That way, by looking at any player we get information about the team. In addition to that, the robot's previous position is taken into account by including its velocity (change in position in that given time step) into the state. In fact, velocity and previous position can be used interchangeably here, because both can be calculated given the current position of a player. In that sense, including the player's velocity adds information about tendency of individual movement (e.g. wingers that tend to move on the upper half of the field, and never go too deep inside it).

All of the previous is done with a grid approach, where the x and y coordinates from the vision packages are transformed into a cell in the grid that handles noisy readings. Given the memory requirements for this system, a 3 by 2 grid was chosen.

In summary, a state is formed by three components: team centroid, position, and velocity. It has in total six dimensions: team centroid x and y coordinates, position x and y coordinates, and velocity (i.e. previous position) on x and y directions. The number of possible states grows exponentially as each component can take any of the possible cells of the grid.

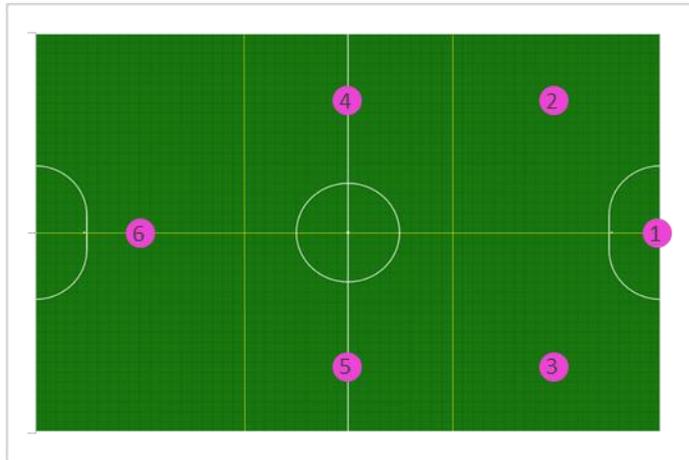

Figure 1. Sample reading from vision package

| team centroid (x,y) | (1,1) |
|---|---|
| position(x,y) | (2,1) |
| velocity(x,y) | (0,0) |
| state | 1<(1,1),(2,1),(0,0)> |

Figure 2. Sample state for player 1

| Initial formation |
|---|
| 1<(1,1),(2,1),(0,0)> |
| 2<(1,1),(2,1),(0,0)> |
| 3<(1,1),(2,0),(0,0)> |
| 4<(1,1),(1,1),(0,0)> |
| 5<(1,1),(1,0),(0,0)> |
| 6<(1,1),(0,1),(0,0)> |

Figure 3. Sample corresponding formation

### 3.2. Probability distributions

The next elements to define for a Markov chain are its initial state and state transition probability distributions. For this project, an initial state probability table is not needed because the states of the players in the initial formation are deterministic. The game starts with a vision package that gives information about each robot position and the team centroid. To finalize each player's initial state, the assumption is made that at the initial state all velocities are 0, meaning that at the beginning of the game they were not moving or that their previous position does not matter.

In contrast, defining and learning the state transition probability distribution is the key element for this project. The way the probability is calculated is based on a count system thus, each time a state transition is seen during the learning phase, its count will be increased by one. That way, the probability of having a specific transition simply corresponds to the count of the desired state divided by the total count of corresponding relevant states.

### 3.3. The transition table

The system keeps track of the state transition counts by the use of a table of size $\theta((\#cells\ in\ the\ grid^3)^2)$, where $\#cells\ in\ the\ grid = 3\ x\ 2 = 6$ for this project. For any element in the transition table, its row represents the FROM state whereas its column represents the TO state.

The table is highly efficient, both in memory and running time requirements. It is stored as a vector matrix, which is running time efficient by itself. However, to make it memory efficient, the relation between state transitions and cells in the table must be one to one. In order to do so, every state is transformed into a unique index in the table similar to how a base transformation is done. The formula is as follows:

$$\begin{aligned}Index\ in\ table &= \left(v_y * (g_l{}^0 * g_w{}^0)\right) + \left(v_x * (g_l{}^0 * g_w{}^1)\right) + \left(p_y * (g_l{}^1 * g_w{}^1)\right) \\ &+ \left(p_x * (g_l{}^1 * g_w{}^2)\right) + \left(t_y * (g_l{}^2 * g_w{}^2)\right) + \left(t_x * (g_l{}^2 * g_w{}^3)\right) \\ &= (v_y * 1) + (v_x * 2) + (p_y * 6) + (p_x * 12) + (t_y * 36) + (t_x * 72)\end{aligned}$$

$$\begin{aligned}where\ v &= velocity \\ p &= position \\ t &= team\ centroid \\ g_l &= grid\ length \\ g_w &= grid\ width\end{aligned}$$

For visualization purposes, "statesList.cpp" lists all the possible states in its decomposed mode and its corresponding index.

## 4. Learning

### 4.1. Data

At the beginning of this project, the goal was set to train the system with logs from previous official games at the RoboCup SSL. A second alternative was to use grSim[10] to generate offence plays and extract vision packages. However, interpreting vision packages is a different problem that goes beyond the scope of this project. For this reason, dummy data was created specifically to train and test the system.

Three different offence pre-designed plays were created: one regular pass-shoot play, and two corner plays, one from each corner (See Appendix 1). These plays consisted of three steps each, which means that each play has a sequence of three different formations that is unique to it. Short plays are ideal since the purpose of the system is to provide with responsiveness, not complex play modeling.

From these artificially created plays, we get a pool of state transitions. Given that each formation has 6 player states, and that each play consists of three formations (meaning two transitions between formations), it is found that:

$$state\ transitions_{data}$$
$$= \#\ states\ per\ formation * \#\ formation\ transitions\ per\ play * \#\ plays$$
$$= 6 * 2 * 3 = 36$$

The calculation assumes that players move in distinct manners during the predesigned plays, and so they generate distinct state transitions.

We also know that the total space of possible state transitions is:

$$states\ transitions_{total} = (\#cells\ in\ the\ grid^3)^2 = 6^6 = 46\ 656$$

Therefore, the dummy data represents only $\frac{1}{1296}$ of the possible state transitions.

To counterbalance overfitting, a random element is added that expands the state transitions space seen while learning. Every time this random element is called, it generates a formation where all players are placed at random positions. This serves two purposes. First, it introduces noise into the system, hence modelling the process more realistically. Second, since the positions are random, when the system transforms the information into states, these will -in theory- be different from the ones in predesigned plays. As a consequence, the pool of states the robots come from and go to grows.

The predesigned plays are hardcoded as vision packages into "packageProducer.cpp", a program that simulates the vision system and delivers a package every 2 seconds. A random number generator decides which package to deliver. If it chooses 1, 2 or 3, the corresponding predesigned play is activated and its three formations are delivered. In the case of 0, only one random formation is delivered. Therefore, 90% of the readings correspond to plays formations and only 10% to noise.

Once again, this is a temporary approach to test the performance of the system. Training with real data should lead to better results that are not dependent on plays, but only on the vision package received at the time.

### 4.2. Training

With a simulation of the vision system, the system is now ready to learn the state transition probability distribution. The program "training.cpp" reads the vision package emitted by "package producer.cpp", handles the data, and records the relevant information in the count table.

The way a vision package is transformed into a formation is as follows:

- Team centroid: The system first reads all the x and y position coordinates and computes the average coordinates. Then it transforms those coordinates into a grid cell and sets that as the team centroid.
- Position: Second, each player's position coordinates are transformed into a grid cell and stored as part of its state.
- Velocity: Finally, the system builds its initial formation by assuming the velocity of every player is 0. After that point, the system calculates velocity by getting the player's position from the previously stored formation and recording the difference with the current position.

Note that while the system stores formations as a whole, it only records transitions from one state to another by looking at a specific player's state in the previous and current formation. Every time a transition is seen, its corresponding value in the table is incremented by one.

The system received a total of 15 000 vision packages corresponding to approximately 10 hours of training.

## 5. Predicting formations

Since the system is trained with plays of three steps, the approach to test it is based on a three step prediction. The program "predictor.cpp" receives a random formation and predicts two successive opponent formations.

### 5.1. Restriction by team centroid

To make a prediction, some restrictions must be set on the transitions states can undergo. One of the most important ones relates to the team centroid, since the fact that every player on the team has the same team centroid is an invariant. Hence, when the system makes a prediction it first calculates the most likely team centroid the team will move to given the current centroid.

The system calculates the most likely centroid by segmenting the transition table into blocks of team centroid transitions. Each team centroid has a block size of $\#cells\ in\ the\ grid^2$ possible states related to it. This means that the FROM team centroid limits the search to rows between

$$Start\ index = \left(t_y * (g_l^2 * g_w^2)\right) + \left(t_x * (g_l^2 * g_w^3)\right)$$
$$End\ index = Start\ index + team\ centroid\ block\ size$$

Next, the columns are grouped into team centroid blocks using the same centroid block size. All the state transition counts on that block are added. The most likely team centroid is the team centroid block with the largest count. The probability associated with the chosen team centroid is

$$P(Team\ centroid) = \frac{Max\ TO\ team\ centroid\ count}{Total\ FROM\ team\ centroid}$$

The method ensures consistency across the predicted formation. It also brings additional benefits to running time because the search of each player's most likely state transition is reduced to look at only a block of $O((\#cells\ in\ the\ grid^2)^2)$ elements.

### 5.2. Probability of formation and confidence of prediction

The next step to assign a new state to each player is looking for the most likely state it will transition to. Given that the FROM state is known (i.e. the row index is known), and that the search was reduced to only one team centroid block (i.e. the column indexes are restricted), space complexity of the search per player is in $O(\#cells\ in\ the\ grid^2)$. The process for calculating the probability of each player's state transition is similar to finding the most likely team centroid, such that the probability of player $p$ transition is

$$P(State\ transition_p) = \frac{Max\ TO\ state\ count}{Total\ FROM\ state}$$

Each new chosen state is saved into a temporal structure with its related probability. Finally, the predicted formation is built with each player's new state. The probability of the formation is

$$P(Formation) = P(Team\ centroid) * \left(\prod_{p=1}^{\#players} P(State\ transition_p)\right)$$

In this work the first formation is read from the vision package and only two other predictions are made to finish predicting a play. Every prediction is based on a previous formation. Therefore, the further in the future a prediction is made, the lower its confidence level since the base formation was initially uncertain. Hence the prediction confidence is calculated as

$$Confidence\ level\ (current\ prediction) \\ = Confidence level\ (Base\ formation) * P(predicted\ formation)$$

## 6. Evaluating the system

### 6.1. Measures

The system is evaluated by measuring the similarity of the predicted formation to the training play formation in two different ways: count and distance.

The count measure counts the total number of players in the formation whose predicted state does not correspond with the state in the predesigned play formation. It takes into account all of the components of the state, and hence is more strict in evaluating predictions. It requires that the robot is in the correct position, has the correct velocity and that the rest of the team behaves accordingly. The range per predicted formation of this measure is $[0,6]$ since there are up to six players whose state could be wrongly predicted.

The distance measure adds the displacement of each player's predicted position away from the position in predesigned play formation. It has a wider range of $[0, (6 * \max displacement)]$ since the distances for a displaced robot may grow substantially as the grid becomes finer. The main disadvantage of this measure is that it does not take into account neither the team centroid nor the velocity, thus accepting senseless states as correct. Predictions that violate the team centroid invariant, or that have impossible velocities, are not penalized.

Using the two types of similarity measures, the system compares the initial formation (read from the vision package) to the initial formations of each of the predesigned plays. For each type of measure, it selects the play that scored the lowest and compare the formation predictions of step 2 and 3 to the corresponding formations of the selected play.

### 6.2. Results and analysis

The results for the testing phase can be divided into two: similarity measures performance and confidence level performance.

With regards to the similarity measures, the final results are as follows:

Number of tests: 5000

|  | Average measure per play | Average measure per formation |
| --- | --- | --- |
| Count similarity measure | 16.6826 | 5.5608 |
| Distance similarity measure | 21.9680 | 7.3226 |

The data shows that on average, about 5 of the predicted robot states differ from the states they should have according to the selected predesigned play. This means that at every formation, only one of the player's states was predicted correctly.

On the other hand, the distance similarity measure reveals that the total displacement between the robots' predicted positions and their predesigned play positions is only 7 grid cells. The chosen grid for this project consists of 6 cells which is in the same order of magnitude as the 7 cell total displacement. It is easy to imagine two scenarios: either each player's position was displaced by a one or two cells only, or in an outlier's case, a few robots have a predicted position that corresponds to a large displacement. Further analysis of this measure is needed to determine which of the cases occurs more often.

Gathering the results from both measures, it is concluded that the system does not make accurate predictions under the circumstances tested. When the whole state is taken into account, 5 out of 6 players are wrongly modelled. When just the position is taken into account, it is noted that the displacement is probably the main reason for the wrong state predictions.

As an indicator of the validity of the similarity measures, the following graph shows that the average for both measures stabilizes around the $800^{th}$ test.

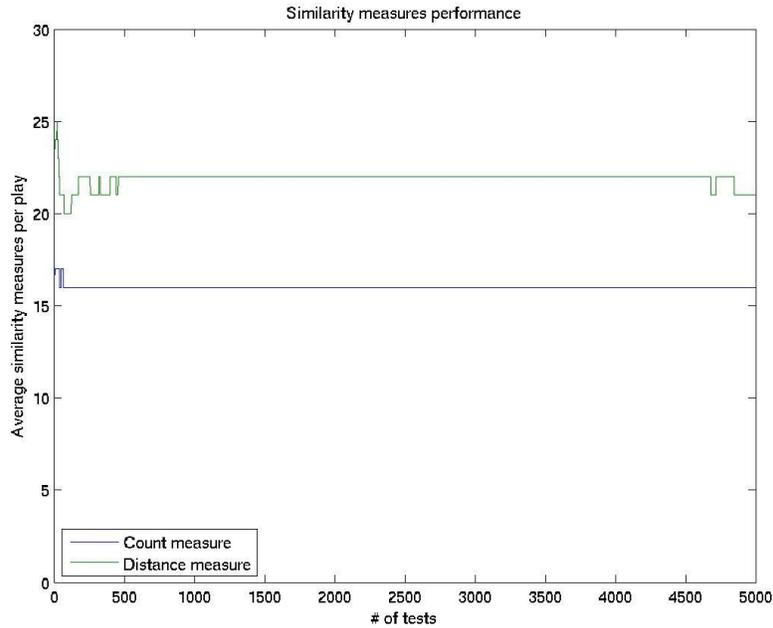

*Figure 4. Average similarity measures*

The average confidence level for the first and second predicted formations are also calculated. The following graph shows that there is a rapid decrease in confidence level as the predictions are made further in the future. However, since the goal of the system is to be responsive, it should only be concerned with improving the confidence level for the first couple of predictions.

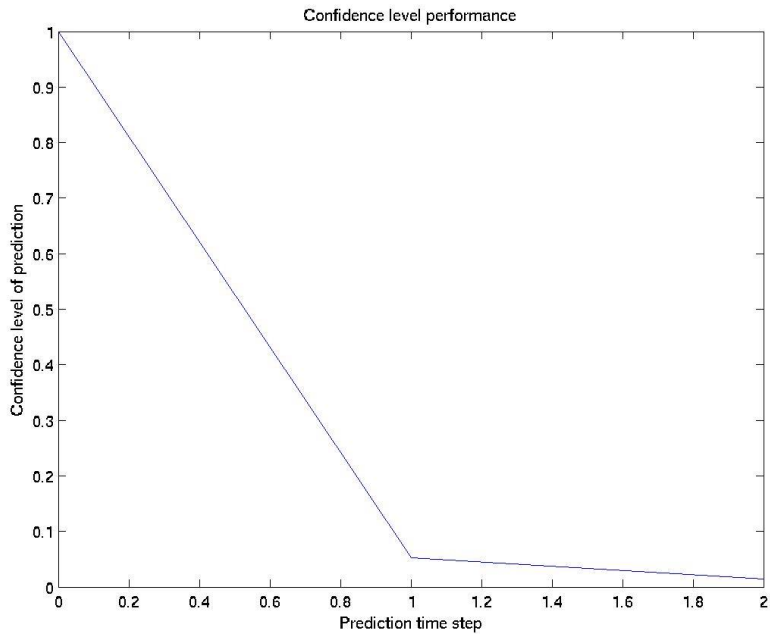

*Figure 5. Confidence level decrease*

## 7. Discussion

The similarity measures raise some interesting questions that should be further analysed. For example, it would be interesting to see how the size of the grid affects the distance measure. It would also be valuable to investigate if the displacement is widespread to most of the players, of if it is only some that bring the number to high values. Finally, it is important to investigate how the system behaves when trained with real data and not only a small pool of predesigned plays.

On a different note, it is worth mentioning that the system produces predictions with extreme levels of confidence; that is, it is either a very high confidence level, or a really low one. This is due to the poor learning procedure implemented which prevents the transition table to be populated evenly. Since the states considered in pre-designed plays are clustered in the table, the initial formation is "pulled" towards these sections creating high confidence values. However, the team centroid restriction may make it impossible to reach the populated sections of the table, hence producing low confidence values.

Still, the system produces sufficient information that can be read in many different manners to improve a team's defence. Each component of a state opens opportunities for defence tactics to improve. For example, the team centroid brings insight on tendencies about how the team evolved its defence-offence balance. The velocity corresponds to the change in position, and so it serves the purpose of blocking passes and moves.

### 7.1. Improvements and further work

One of the major improvements for this system is to restrict even further the pool of states to which one state can transition to. By restricting it by velocity (meaning previous position and current position have to be correct) as well as team centroid, the system predictions will be more meaningful. However, there is a risky trade-off since restricting the pool of states might lead to over-fitting. Since this project had limited data from the beginning, a finer rejection mechanism was not implemented.

A second, simpler improvement is to allow rotated fields. At the moment, this system only interprets a play as offensive when the opponent team tries to advance to the right. However, a simple function could interpret the vision package and rotate it as necessary if the team faces left. In fact, Peter Carr and Manuela Veloso incorporated simple rotation functions in their systems that could be incorporated on to this project.

Regarding the implementation on real robots, the methods for rapid search provided decrease significantly the work need to be done to predict a future formation. Additionally, if dynamic programming were to be implemented to calculate the most likely centroid transitions, the running time would be directly improved. Furthermore, even though a grid approach comes with challenges about exponential growth, the real impact it has on running time will not be significant during the games. This project showed that training the system is the most crucial and time consuming part.

The final improvement arises from the difficulties for getting a hold of training and testing data. As a basic rule, training and testing sets must consist of different data. However, due to the constraints of the project, this rule was broken. Real time data from the SSL vision system, logs from previous games[11], and simulations will provide with the variety needed for the correct evaluation of the system and significantly increase its the usefulness.

# Appendix 1: Training data

Play 1 Formation 1

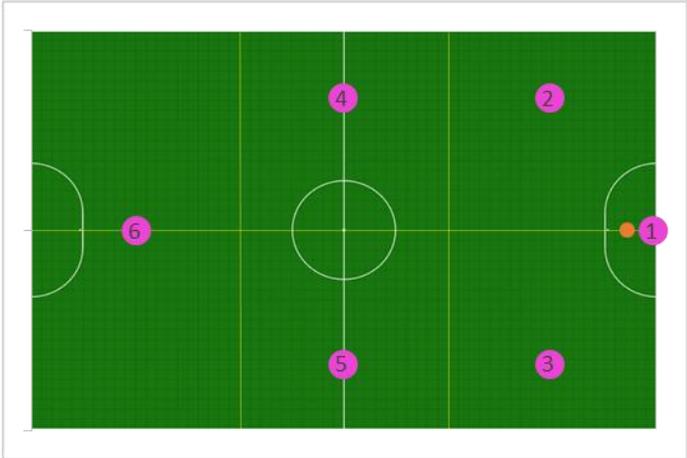

Play 1 Formation 2

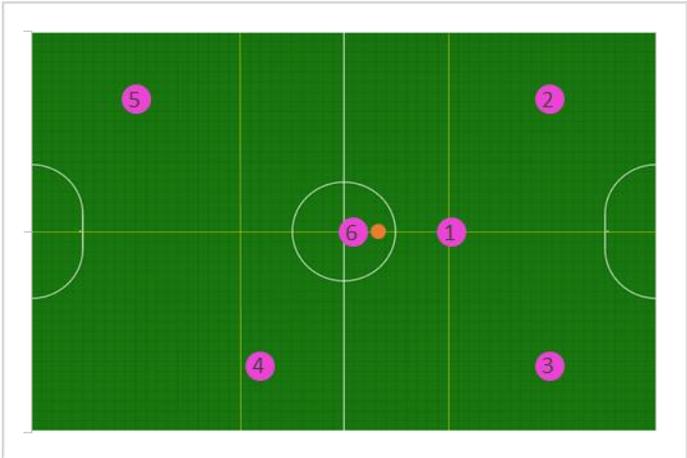

Play 1 Formation 3

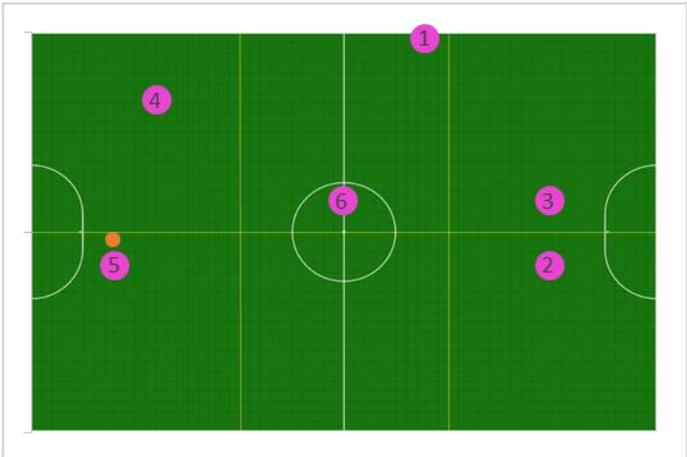

Play 2 Formation 1

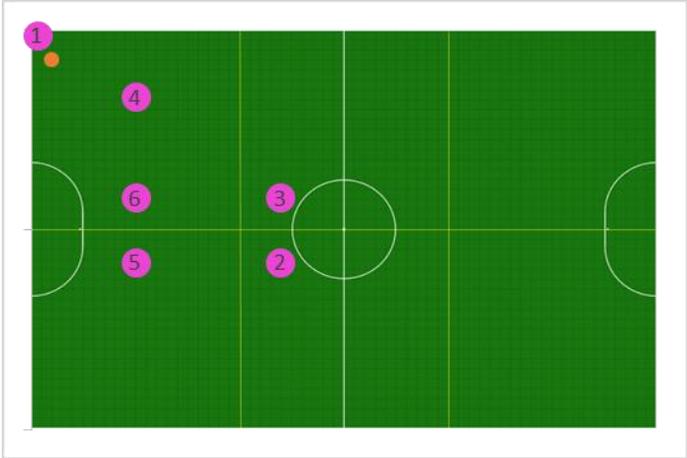

Play 2 Formation 2

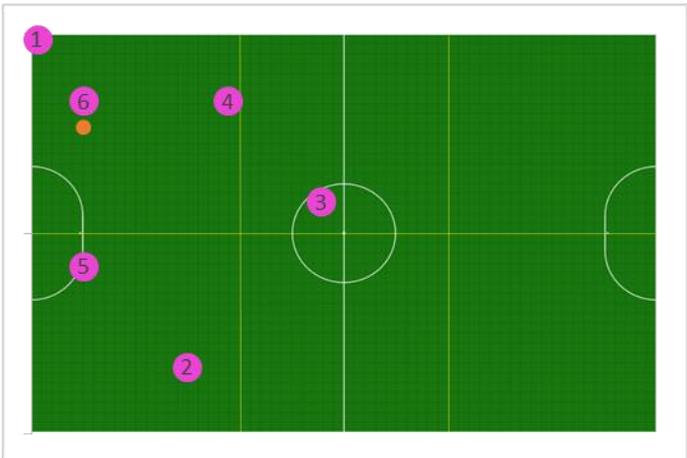

Play 2 Formation 3

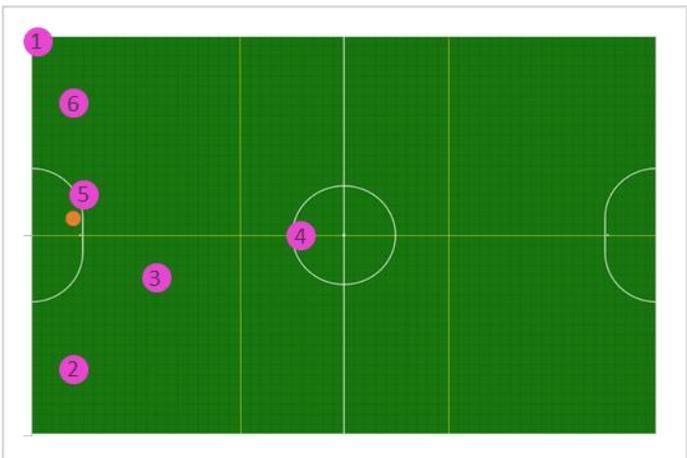

Play 3 Formation 1

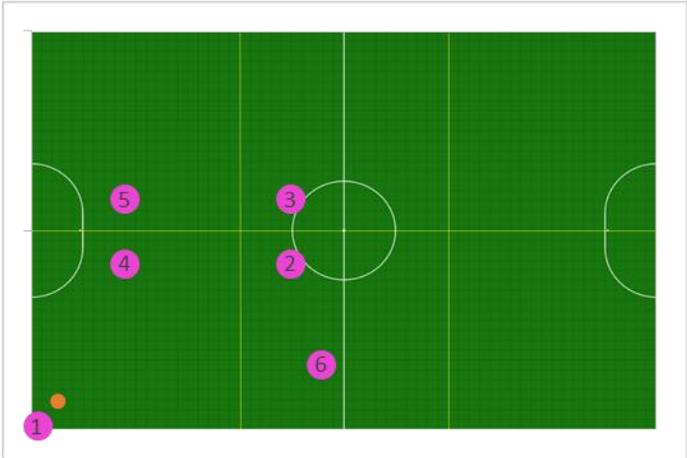

Play 3 Formation 2

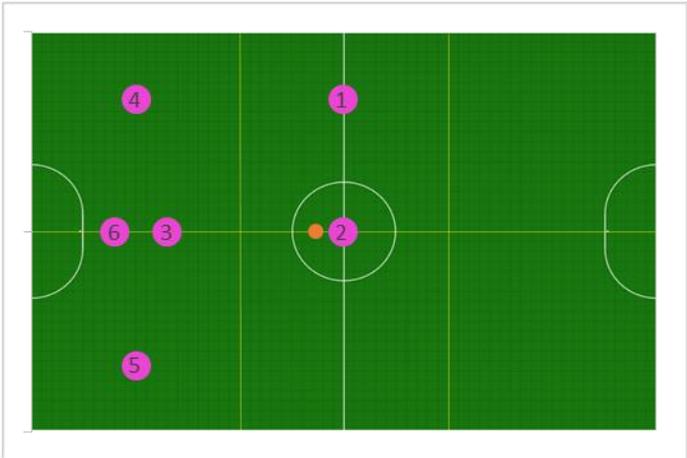

Play 3 Formation 3

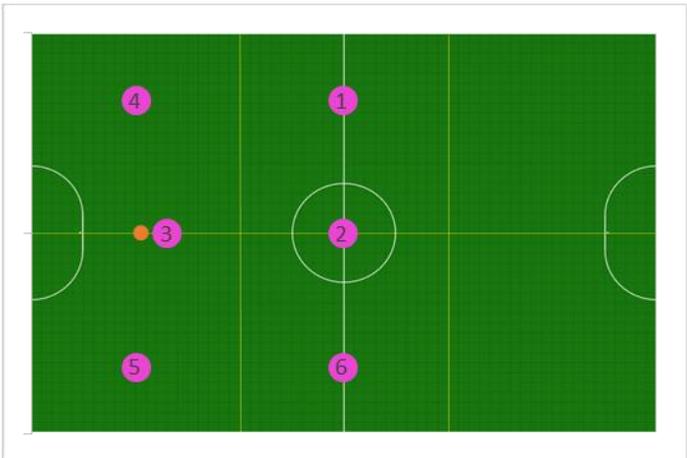

## 9. References

[1] Sahota, M. K., Mackworth, A. K., Kingdon, S. J., & Barman, R. A. (1995, October). Real-time control of soccer-playing robots using off-board vision: the dynamite testbed. In *Systems, Man and Cybernetics, 1995. Intelligent Systems for the 21st Century., IEEE International Conference on* (Vol. 4, pp. 3690-3693). IEEE. Retrieved from http://ieeexplore.ieee.org/xpl/articleDetails.jsp?arnumber=538361

[2] Bruce, J., Bowling, M., Browning, B., & Veloso, M. (2003, September). Multi-robot team response to a multi-robot opponent team. In *Robotics and Automation, 2003. Proceedings. ICRA'03. IEEE International Conference on*(Vol. 2, pp. 2281-2286). IEEE. Retrieved from http://www.cs.cmu.edu/~mmv/papers/03icra-cmdragons02.pdf

[3] Bowling, M. H., Browning, B., & Veloso, M. M. (2004, January). Plays as Effective Multiagent Plans Enabling Opponent-Adaptive Play Selection. In*ICAPS* (pp. 376-383). Retrieved from http://www.aaai.org/Papers/ICAPS/2004/ICAPS04-044.pdf

[4] Biswas, J., Mendoza, J. P., Zhu, D., Choi, B., Klee, S., & Veloso, M. (2014, May). Opponent-driven planning and execution for pass, attack, and defense in a multi-robot soccer team. In *Proceedings of the 2014 international conference on Autonomous agents and multi-agent systems* (pp. 493-500). International Foundation for Autonomous Agents and Multiagent Systems. Retrieve from http://aamas2014.lip6.fr/proceedings/aamas/p493.pdf

[5] Erdogan, C., & Veloso, M. (2011, July). Action selection via learning behavior patterns in multi-robot domains. In *Proceedings of the Twenty-Second international joint conference on Artificial Intelligence-Volume Volume One* (pp. 192-197). AAAI Press. Retrieved from http://ijcai.org/papers11/Papers/IJCAI11-043.pdf

[6] Trevizan, F. W., & Veloso, M. M. (2010, November). Learning Opponent's Strategies In the RoboCup Small Size League. In *9th International Conference on Autonomous Agents and Multi-Agent Systems. Springer*. Retrieved from http://www.cs.cmu.edu/~mmv/papers/10aamasw-sslearning.pdf

[7] Bialkowski, A., Lucey, P., Carr, P., Denman, S., Matthews, I., & Sridharan, S. (2013, June). Recognising team activities from noisy data. In *Computer Vision and Pattern Recognition Workshops (CVPRW), 2013 IEEE Conference on* (pp. 984-990). IEEE. Retrieved from: http://ieeexplore.ieee.org/stamp/stamp.jsp?tp=&arnumber=6595989

[8] Chakraborty, D., Agmon, N., Stone, Peter, Targeted Opponent Modeling of Memory-Bounded Agents. In *Proceedings of the Adaptive Learning Agents Workshop (ALA)*, May 2013. Retrieved from http://www.cs.utexas.edu/~pstone/Papers/bib2html-links/ALA13-chakrado.pdf

[9] Butler, S., & Demiris, Y. (2009). Predicting the movements of robot teams using generative models. In *Distributed Autonomous Robotic Systems 8* (pp. 533-542). Springer Berlin Heidelberg. Retrieved from http://link.springer.com/chapter/10.1007/978-3-642-00644-9_47

[10] Monajjemi, V., Koochakzadeh, A., & Ghidary, S. S. (2012). grsim–robocup small size robot soccer simulator. In *RoboCup 2011: Robot Soccer World Cup XV* (pp. 450-460). Springer Berlin Heidelberg. Retrieved from: http://link.springer.com/chapter/10.1007/978-3-642-32060-6_38

[11] Real game logs can be found here http://robocupssl.cpe.ku.ac.th/gamelogs